\documentclass[10pt, a4paper]{article}

\usepackage[final]{lrec2026} 

\usepackage{graphicx}
\usepackage{tabularx}
\usepackage{soul}
\usepackage{float}
\usepackage{comment}
\usepackage{xcolor}
\usepackage{hyperref}
 \definecolor{darkblue}{rgb}{0, 0, 0.5}
  \hypersetup{colorlinks=true, citecolor=darkblue, linkcolor=darkblue, urlcolor=darkblue}
\usepackage{xstring}
\usepackage{color}
\usepackage[table,xcdraw]{xcolor}
\usepackage{todonotes}

\title{The Growing Gains and Pains of Iterative Web Corpora Crawling:\\Insights from South Slavic CLASSLA-web 2.0 Corpora}
\name{Taja Kuzman Pungeršek$^{\ast}$, Peter Rupnik$^{\ast}$, \\ {\bf \large Vít Suchomel$^{\S}$, Nikola Ljubešić$^{\ast}$$^{\dagger}$$^{\ddagger}$}} 

\address{$^{\ast}$Jožef Stefan Institute; $^{\dagger}$Faculty of Computer and Information Science, University of Ljubljana;\\ $^{\ddagger}$Institute of Contemporary History; $^{\S}$Natural Language Processing Centre, Masaryk University\\
         $^{\ast}$$^{\dagger}$$^{\ddagger}$Ljubljana, Slovenia; $^{\S}$Brno, Czech Republic \\
         \{taja.kuzman, peter.rupnik, nikola.ljubesic\}@ijs.si; $^{\S}$xsuchom2@fi.muni.cz\\}

\abstract{
Crawling national top-level domains has proven to be highly effective for collecting texts in less-resourced languages. This approach has been recently used for South Slavic languages and resulted in the largest general corpora for this language group: the CLASSLA-web 1.0 corpora. Building on this success, we established a continuous crawling infrastructure for iterative national top-level domain crawling across South Slavic and related webs. We present the first outcome of this crawling infrastructure -- the CLASSLA-web 2.0 corpus collection, with substantially larger web corpora containing 17.0 billion words in 38.1 million texts in seven languages: Bosnian, Bulgarian, Croatian, Macedonian, Montenegrin, Serbian, and Slovenian. In addition to genre categories, the new version is also automatically annotated with topic labels. Comparing CLASSLA-web 2.0 with its predecessor reveals that only one-fifth of the texts overlap, showing that re-crawling after just two years yields largely new content. However, while the new web crawls bring growing gains, we also notice growing pains -- a manual inspection of top domains reveals a visible degradation of web content, as machine-generated sites now contribute a significant portion of texts.
 \\ \newline  }

\begin{document}

\maketitleabstract


\section{Introduction}

\begin{figure*}[!ht]
\begin{center}
\includegraphics[width=\textwidth]{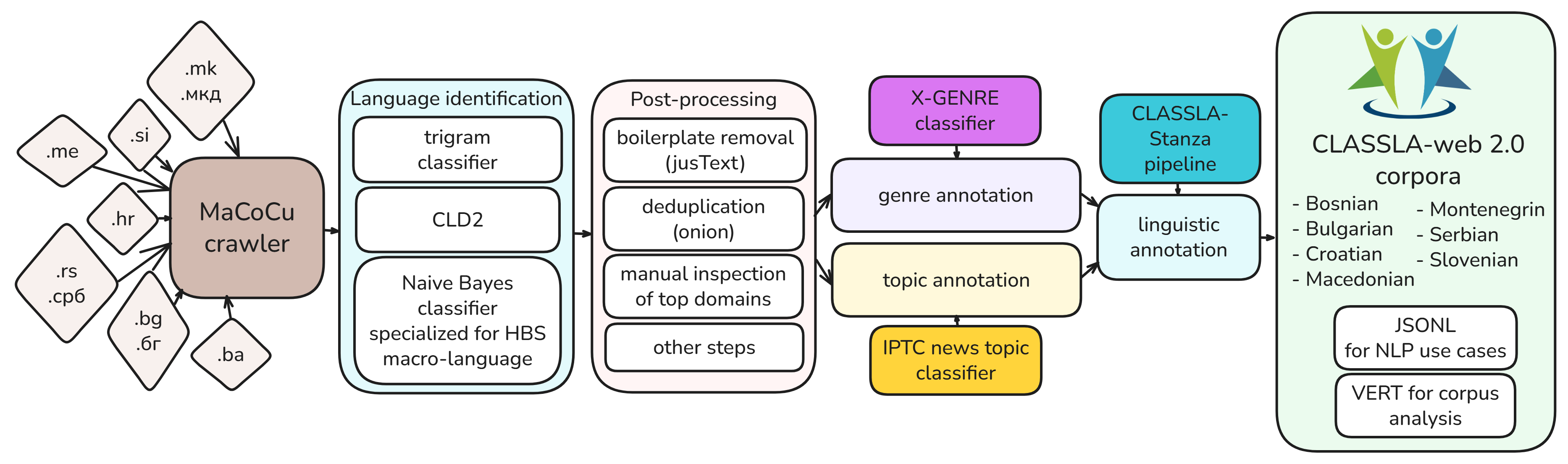}
\caption{The CLASSLA-web construction pipeline (detailed in Section \ref{sec:pipeline}), consisting of several key steps: web crawling, language identification using multiple tools, post-processing to ensure high-quality corpus content, automatic genre and topic annotation, and linguistic annotation.}
\label{fig:classla-web-pipeline}
\end{center}
\end{figure*}

Building large, high-quality corpora for less-resourced languages remains a central challenge for the natural language processing field. Crawling top-level domains (TLD) has been shown to be a very efficient method to collect large numbers of texts for less-resourced languages. For South Slavic languages, this approach recently resulted in the largest general corpora for this language group -- the CLASSLA-web 1.0 corpora \citep{ljubevsic2024classlaweb}. The corpus collection has been widely used for pre-training BERT-like and decoder-only large language models \citep{ljubesic-etal-2024-language,vrevs2024generative}, for building datasets for downstream natural language processing (NLP) tasks~\citep{kuzman2025newstopic,Kuzman_Pavleska_Rupnik_Cigoj_2024}, and has been shown to be heavily consulted by linguists~\citep{erjavec2024clarin,ljubevsic2024classlaexpress}.

Motivated by these positive results, we set up an ongoing crawling infrastructure dedicated to iterative national top-level domain (TLD) crawling of South Slavic and other webs. In this paper, we present the first outcome of the crawling infrastructure -- the CLASSLA-web 2.0 corpus collection, a new release of web corpora for seven South Slavic languages. As the crawling pipeline methodologically mirrors the crawling pipeline from the MaCoCu project \citep{banon2022macocu}, with which the CLASSLA-web 1.0 South Slavic corpus collection was produced, this enables us a comparison over time and provides insights into the evolution of the web in the two years between the two crawls.

Our study makes the following contributions. First, we release a substantially larger web corpus collection containing 17.0 billion words in 38.1 million texts across Bosnian, Bulgarian, Croatian, Macedonian, Montenegrin, Serbian, and Slovenian (presented in Section \ref{sec:classla-web-2.0}). The CLASSLA-web 2.0 corpora can be downloaded in JSONL and VERT formats from the CLARIN.SI repository \citeplanguageresource{classlaweb2-repository}\footnote{The CLASSLA-web 2.0 corpora are freely accessible in the CLARIN.SI repository at \url{http://hdl.handle.net/11356/2079}.} and queried through the CLARIN.SI concordancers\footnote{\url{https://www.clarin.si/ske/}, the links to each of the corpora are provided on the website \url{https://clarinsi.github.io/classla-web/}.}. Each text is linguistically annotated and labelled with genre and topic categories. Leveraging these annotation layers, we provide a detailed view of the distribution of genres and news topics across South Slavic webs (Section \ref{sec:genre-topic-analysis}).

Second, we perform a comparison between the CLASSLA-web 1.0 and CLASSLA-web 2.0 corpora (Section \ref{sec:version-comparison}) showing that repeating TLD crawls after only two years yields much larger corpora with predominantly new content: on average, 82\% of texts in CLASSLA-web 2.0 do not occur in CLASSLA-web 1.0. Analysing the correlation between the URL overlap and content overlap between two CLASSLA-web versions, we propose an approximate function for estimating content overlap via a weighted linear regression (Section \ref{sec:url-overlap-calculation}). Third, we quantify web evolution in terms of change of sizes (Section \ref{sec:size-comparison}), and web content quality (Section \ref{sec:degradation-of-content}). We report a visible degradation of web content between versions, with ``bad'' (e.g., machine-generated or machine-translated) domains now contributing a much larger share of crawled material, underscoring the necessity of manual inspection of web domains that contribute the largest amounts of text.


\section{Related Work}
\label{sec:related-work}

The practice of building large web corpora for European languages was established with the WaCky initiative~\citep{baroni2009wacky}. The first web corpora of South Slavic languages were compiled for Slovenian (slWaC) and Croatian (hrWaC,~\citealp{ljubevsic2011hrwac,erjavec2015slwac}), followed by Bosnian (bsWaC) and Serbian (srWaC,~\citealp{ljubesic-klubicka-2014-bs}). After a seven-year gap, the MaCoCu project~\citep{banon2022macocu} resumed crawling activities for South Slavic languages and other less-resourced European languages. The monolingual web corpora for South Slavic languages that were produced in the MaCoCu project have later been additionally post-processed and enriched with linguistic and genre annotation, and published as CLASSLA-web 1.0 corpus collection~\citep{ljubevsic2024classlaweb}.
The collection spans Bosnian~\citelanguageresource{classla-bs}, Bulgarian~\citelanguageresource{classla-bg}, Montenegrin~\citelanguageresource{classla-cnr}, Croatian~\citelanguageresource{classla-hr}, Macedonian~\citelanguageresource{classla-mk}, Slovenian~\citelanguageresource{classla-sl}, and Serbian~\citelanguageresource{classla-sr} web corpora.
The CLASSLA-web corpora have received considerable attention in the South Slavic space, both from the NLP community \citep{ljubesic-etal-2024-language,vrevs2024generative,kuzman2025newstopic,Kuzman_Pavleska_Rupnik_Cigoj_2024} and from linguists and language teachers \citep{erjavec2024clarin,ljubevsic2024classlaexpress}. 

In parallel, numerous other web-based datasets have emerged from the Common Crawl and Internet Archive data collections, e.g., the cc100 dataset~\citep{conneau-etal-2020-unsupervised}, the mC4 dataset~\citep{xue-etal-2021-mt5}, the OSCAR dataset~\citep{suarez2019asynchronous}, and most recent, the HPLT \citep{de-gibert-etal-2024-new-massive} and FineWeb2 datasets \citep{penedo2025fineweb2}. While these multilingual collections are highly useful for multilingual language modelling tasks, they suffer, inter alia, from lower quality of content, namely, cc100 and mc4 datasets \citep{van2024language}; or an unexplainable small amount of content in certain South Slavic languages, namely in OSCAR\footnote{\url{https://oscar-project.github.io/documentation/versions/oscar-2301/}}, mC4 dataset~\citep{xue-etal-2021-mt5}, the HPLT 3.0 datasets \citep{de-gibert-etal-2024-new-massive}, and the FineWeb2 dataset \citep{penedo2025fineweb2}. More specifically, the most recent HPLT 3.0 \citep{de-gibert-etal-2024-new-massive} and FineWeb2 \citep{penedo2025fineweb2} datasets are missing Montenegrin language. Moreover, they report unusual sizes of Serbian and Bosnian corpora where the Bosnian web corpus is significantly larger than the Serbian one, which does not correlate with the number of language speakers or web corpus sizes reported for the CLASSLA-web corpora \citep{ljubevsic2024classlaweb}. 
It seems that these corpora that use ``one pipeline to scale them all'' suffer from language identification issues related to the Serbian, Bosnian and Montenegrin languages. 
Furthermore, since the introduction of large language models that make text generation in various languages easily available, there is a growing problem on the web that more and more content is automatically generated, as we will show in Section \ref{sec:degradation-of-content}. It is thus crucial that manual inspection of web corpora content by native speakers is included in the corpora preparation to filter out as much bad content as possible, and to assure high quality of data. Our work addresses these gaps via targeted TLD crawling, specialised language discrimination between the mutually intelligible Croatian, Serbian, Bosnian and Montenegrin languages, and manual domain validation.

\section{CLASSLA-web Construction Pipeline}
\label{sec:pipeline}

The goal of the crawling pipeline is to produce large, clean, and richly annotated web corpora. As shown in Figure \ref{fig:classla-web-pipeline}, the process consists of 1) crawling based on top-level national domains and post-processing, 2) automatic text annotation with genre and topic information, and 3) linguistic annotation. In this section, we briefly present each step. For more details, refer also to the description of the CLASSLA-web crawling pipeline in \citet{ljubevsic2024classlaweb} and to the website that presents the CLASSLA-web collections (\url{https://clarinsi.github.io/classla-web/}).

\paragraph{Web Crawling Pipeline} Following the methodology set up in the MaCoCu project \citep{banon2022macocu}, we crawl national top-level domains (e.g., \texttt{.si} for Slovenian), as well as connected generic domains (\texttt{.com} and others). Crawling is performed with the MaCoCu crawler\footnote{\url{https://github.com/macocu/MaCoCu-crawler}}, which is based on the SpiderLing crawler~\citep{suchomel2012efficient} and designed for an efficient large-scale web text collection.

\paragraph{Language Identification} Two language identification tools are used at the document and paragraph levels, namely a trigram classifier~\citep{lui2014accurate} and the Compact Language Detector 2 (CLD2, \citealplanguageresource{sites2013compact})\footnote{\url{https://github.com/CLD2Owners/cld2}}. 
Language identification between Bosnian, Croatian, Montenegrin, and Serbian, comprising the HBS macro-language, has been shown to be particularly challenging, as these South Slavic languages exhibit a significant level of mutual intelligibility \citep{rupnik2023benchic}. We directly assign texts that come from national top-level domains to the corresponding web corpus, e.g., web domains from \texttt{.hr} are included in the Croatian web corpus. For generic top-level domains, we use an additional language identification tool specialised for distinguishing between Bosnian, Croatian, Montenegrin and Serbian, namely a Naive Bayes classifier trained on language-specific wordlists extracted from national TLDs~\citep{rupnik2023benchic}. During the development of CLASSLA-web 2.0 corpora, we also explored language identification through zero-shot prompting with large language models, but the specialised classifier proved to be more reliable for HBS disambiguation.

\paragraph{Post-Processing} After crawling, we remove the boilerplate with the jusText\footnote{\url{http://corpus.tools/wiki/Justext}} tool~\citep{pomikalek2011removing} and near-duplicates with the onion\footnote{\url{http://corpus.tools/wiki/Onion}} tool~\citep{pomikalek2011removing}, and discard exact duplicates. We discard texts that are not written in the target language, repair encoding, clean unwanted HTML elements, and filter overly short documents (comprising only short paragraphs with less than 70 characters or consisting of less than 75 words).

\paragraph{Manual Inspection of Top Domains} We perform a manual validation of web domains that provide the most texts in each corpus. We remove texts coming from domains that are manually identified to predominantly contain machine translation, automatically generated text, or severe encoding issues. As shown later (Section~\ref{sec:degradation-of-content}), this step has become crucial for ensuring the high quality of the web corpus. 

\paragraph{Genre Annotation} We use the multilingual X-GENRE classifier 
\citeplanguageresource{x-genre-classifier, x-genre-classifier-huggingface} to automatically assign to each text one of the following labels: \textit{Information/Explanation}, \textit{Instruction}, \textit{News}, \textit{Legal}, \textit{Promotion}, \textit{Opinion/Argumentation}, \textit{Prose/Lyrical}, \textit{Forum}, or \textit{Other} (for a detailed description of labels, refer to~\citet{kuzman2023automatic}). Based on a multilingual AGILE genre classification benchmark\footnote{\url{https://github.com/TajaKuzman/AGILE-Automatic-Genre-Identification-Benchmark}}, the model yields high performance across European languages (micro-F1 and macro-F1 of 0.85) and particularly high scores on South Slavic languages, both in Latin and Cyrillic scripts, namely macro-F1 of 0.89 for Croatian, 0.91 for Macedonian, and 0.94 for Slovenian. 
To ensure high annotation precision, we additionally use the label \textit{Mix} for instances where no single label exceeds a probability of 0.8. This threshold, identified through manual analysis, effectively isolates documents containing multiple genre categories.

\paragraph{Topic Annotation} In CLASSLA-web 2.0 corpora, a new annotation layer is introduced -- in addition to genres, texts are annotated with topics to provide an additional insight into the corpora content. We use a multilingual news topic classifier 
\citeplanguageresource{iptc_model_huggingface} covering 17 top level labels from the IPTC Media Topic NewsCodes schema (e.g., \textit{Politics}, \textit{Science and Technology}, \textit{Sport} -- for a detailed description of labels, refer to~\citet{kuzman2025newstopic}). On a manually-annotated test set in Croatian, Slovenian, Catalan, and Greek, the model achieves macro-F1 of 0.75 and micro-F1 of 0.73. For instances assigned with class probabilities below 0.6, we use the \textit{Mix} label. As with genre annotation, the confidence threshold was identified based on manual analysis. We should note that the classifier has been evaluated for now only on news texts, as it was primarily intended for news topic classification. Thus, in Section~\ref{sec:genre-topic-analysis}, we analyse topics only within the \textit{News} genre.

\paragraph{Linguistic Annotation} All texts are linguistically annotated with the CLASSLA-Stanza pipeline~\citep{ljubevsic2024classla}\footnote{\url{https://pypi.org/project/classla/}}, providing tokenization, lemmatization and morphosyntactic annotation with state-of-the-art accuracy for Slovenian\footnote{Based on the SloBENCH benchmark at \url{https://slobench.cjvt.si/leaderboard/view/11}.}, and a high performance on Croatian, Serbian, Bulgarian, and Macedonian \citep{ljubevsic2024classla}. We use the ``web'' module of the pipeline, which has been shown to perform well on heterogeneous web texts. The Croatian module is applied to Bosnian and Montenegrin. More detailed argumentation behind the module choices is provided in \citet{ljubevsic2024classla}.

\paragraph{Differences in the Pipeline Between Versions 1.0 and 2.0} Although the overall CLASSLA-web pipeline is more or less the same between versions 1.0 and 2.0 of the CLASSLA-web corpora, three improvements were implemented during the development of the CLASSLA-web 2.0 version. First, near-duplicate removal was improved by masking numbers, punctuation, and links during paragraph-level deduplication to improve the efficiency of identifying texts that differ only in numbers. Second, we expanded the manual inspection of the top domains from only Slovenian and Croatian in CLASSLA-web 1.0 to all languages in CLASSLA-web 2.0. Third, we added topic annotation to complement genre labels.

\section{CLASSLA-web 2.0 Corpora}
\label{sec:classla-web-2.0}

In this section, we present the CLASSLA-web 2.0 corpus collection, collected from the South Slavic webs in 2024 and processed following the procedure detailed in Section \ref{sec:pipeline}.

The CLASSLA-web 2.0 collection comprises all seven South Slavic national webs and languages -- Bosnian (bs), Bulgarian (bg), Croatian (hr), Macedonian (mk), Montenegrin (cnr), Serbian (sr), and Slovenian (sl). In total, it consists of 17 billion words and 38 million texts, as reported in Table \ref{tab:corpora-sizes}. The biggest corpus, namely the Bulgarian (CLASSLA-web.bg 2.0) corpus, comprises 6 billion words and 15 million texts. Other large corpora are Serbian (CLASSLA-web.sr 2.0), Croatian (CLASSLA-web.hr 2.0) and Slovenian (CLASSLA-web.sl 2.0) with 2.3 to 3.7 billion words and 5 to 7 million texts. Bosnian (CLASSLA-web.bs 2.0) and Macedonian (CLASSLA-web.mk 2.0) corpora are smaller, comprising 0.7 to 1 billion words and 2.1 to 2.5 million texts. The Montenegrin (CLASSLA-web.cnr 2.0) corpus is the smallest with 0.3 billion words and 0.8 million texts. However, as shown in Section \ref{sec:size-comparison}, the new Montenegrin corpus is twice the size of the Montenegrin corpus included in the CLASSLA-web 1.0 collection and, to the best of our knowledge, represents the largest general Montenegrin corpus available.

\begin{table}[!ht]
\begin{center}
\begin{tabularx}{\columnwidth}{|l|X|X|}
    \hline
        Corpus & Words (billion) & Texts (million)\\ \hline
        CLASSLA-web.bs 2.0 & 1.01 & 2.54 \\ \hline
        CLASSLA-web.bg 2.0 & 5.99 & 14.67 \\ \hline
        CLASSLA-web.hr 2.0 & 3.01 & 5.92 \\ \hline
        CLASSLA-web.mk 2.0 & 0.69 & 2.11 \\ \hline
        CLASSLA-web.cnr 2.0 & 0.29 & 0.79 \\ \hline
        CLASSLA-web.sr 2.0 & 3.71 & 7.24 \\ \hline
        CLASSLA-web.sl 2.0 & 2.31 & 4.79 \\ \hline
        Total & 17.01 & 38.06 \\ \hline
    \end{tabularx}
    \caption{Sizes of CLASSLA-web 2.0 corpora in billion words and million texts for Bosnian (bs), Bulgarian (bg), Croatian (hr), Macedonian (mk), Montenegrin (cnr), Serbian (sr) and Slovenian (sl) corpora.}
    \label{tab:corpora-sizes}
 \end{center}
\end{table}

\subsection{Genre and Topic Distribution}
\label{sec:genre-topic-analysis}

\begin{figure}[!ht]
\begin{center}
\includegraphics[width=\columnwidth]{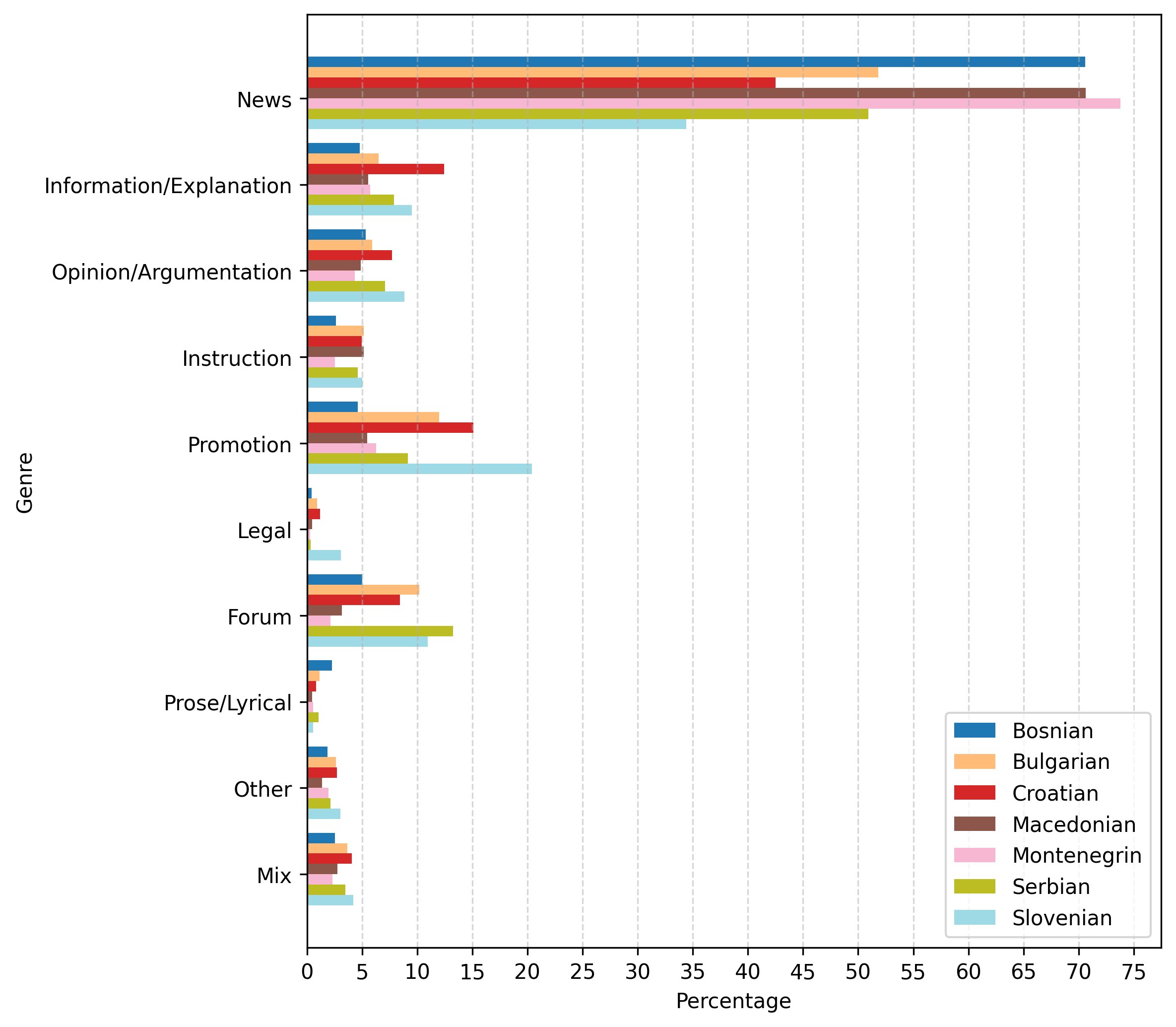}
\caption{Distribution of genre categories across South Slavic CLASSLA-web 2.0 corpora.}
\label{fig:genre-distribution-in-v2}
\end{center}
\end{figure}

Genre and topic labels provide valuable insights into similarities as well as differences between web corpora. As shown in Figure~\ref{fig:genre-distribution-in-v2}, \textit{News} content dominates in all national webs. This genre is especially dominant in Bosnian, Macedonian and Montenegrin corpora, where it exceeds 70\%, whereas it represents 30--40\% of texts in Croatian and Slovenian corpora. As in CLASSLA-web 1.0 corpora \citep{ljubevsic2024classlaweb}, \textit{Promotion} is more prevalent in Slovenian, Croatian and Bulgarian web corpora (12--20\%) than in others. Another genre where there is a visible difference between corpora is \textit{Forum}. Texts from forums seem to be much more frequent in Bulgarian, Serbian, and Slovenian webs (10--13\%) than in Bosnian, Macedonian, and Montenegrin webs (below 5\%).

\begin{figure}[!ht]
\begin{center}
\includegraphics[width=\columnwidth]{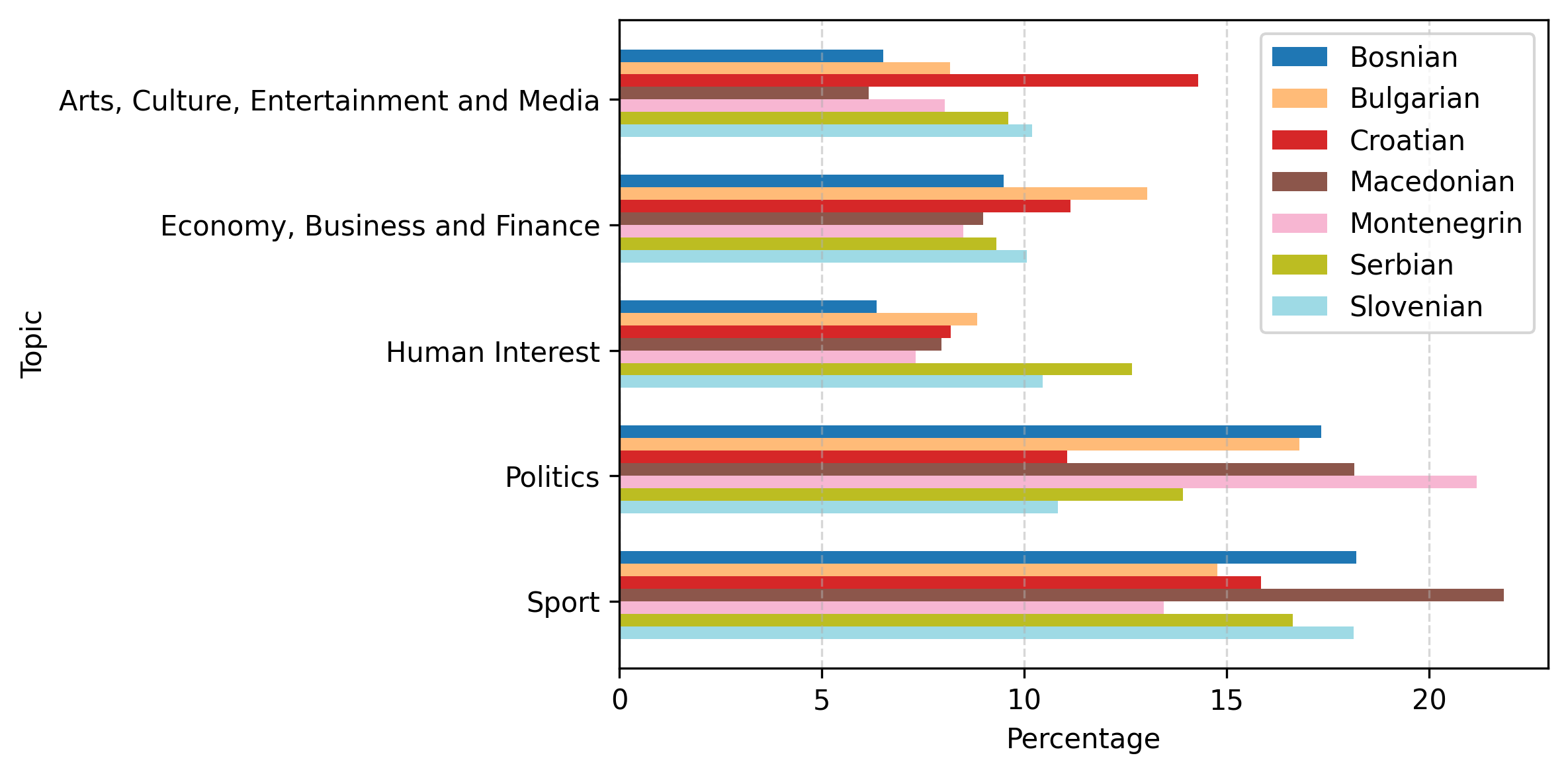}
\caption{Most frequent topics in \textit{News} texts in CLASSLA-web 2.0 corpora. The topics that are represented by 10\% or more texts in at least one corpus are included in the figure.}
\label{fig:most-frequent-topics}
\end{center}
\end{figure}

As \textit{News} is the predominant genre in all web corpora and given that the topic classifier has been primarily validated on news texts, we analyse topics within the \textit{News} genre. In Figure~\ref{fig:most-frequent-topics}, we show the distribution of topics that are discussed in more than 10\% of news texts in at least one corpus. Despite the fact that the news topic schema comprises 17 topic labels, only five categories appear with high enough frequency in the seven corpora, namely ``\textit{Sport}'', ``\textit{Politics}'', ``\textit{Economy, Business and Finance}'', ``\textit{Arts, Culture, Entertainment and Media}'', and ``\textit{Human Interest}'', in this order of average frequency across corpora. These five categories account for approximately 60\% of news texts in each corpus. \textit{Sport} is the most common topic of news texts in the majority of web corpora, and it even represents a fifth of all news in the Macedonian web corpus. In Bulgarian and Montenegrin web corpora, the most frequent topic of news texts is \textit{Politics}, accounting for 20\% of news texts within the Montenegrin corpus. The distribution of the other three most frequent topic categories is relatively consistent across the South Slavic corpora, except in the case of \textit{Arts, Culture, Entertainment and Media} which is notably more prevalent in the Croatian corpus compared to the others, representing almost 15\% of Croatian news texts.

These findings show that the information on genre and topic of web texts -- aside from being very useful for data selection in various research scenarios -- provides valuable insights into the variation of content in South Slavic webs. At the same time, a consistent pattern emerges across all South Slavic corpora: the predominance of news texts on the web, addressing similar topics.

\section{Evolution of the Web between CLASSLA-web 1.0 and CLASSLA-web 2.0}
\label{sec:version-comparison}

To understand how national webs evolve over short time spans, we compare the CLASSLA-web 2.0 corpora with CLASSLA-web 1.0 \citep{ljubevsic2024classlaweb}. Both web corpus collections were compiled using an identical construction pipeline (see Section \ref{sec:pipeline}), with the 1.0 version having been collected from the web two to three years prior to the 2.0 version, specifically in 2021 and 2022. 
In the following sections, we examine the size difference between the two corpora collections (Section \ref{sec:size-comparison}), content overlap (Section \ref{sec:content-overlap}), 
and the presence of automatically generated texts and other low quality content (Section \ref{sec:degradation-of-content}).

\subsection{Size Gains}
\label{sec:size-comparison}

\begin{figure}[!ht]
\begin{center}
\includegraphics[width=\columnwidth]{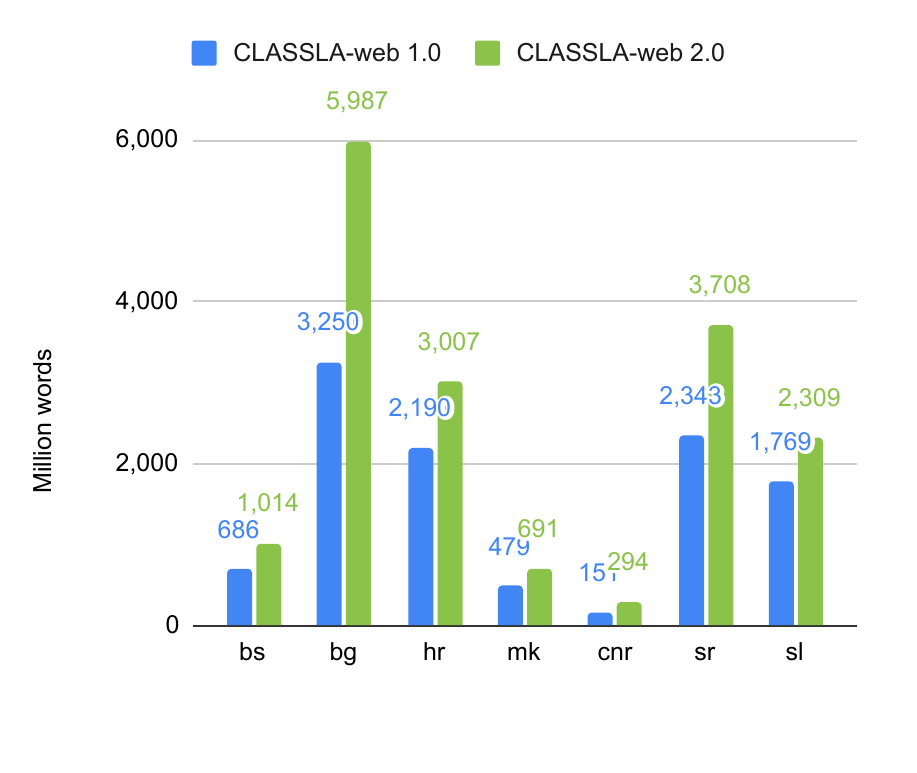}
\caption{Comparison of sizes in millions of words between CLASSLA-web 1.0 and CLASSLA-web 2.0 corpora for Bosnian (bs), Bulgarian (bg), Croatian (hr), Macedonian (mk), Montenegrin (cnr), Serbian (sr) and Slovenian (sl) corpora.}
\label{fig:size-comparison}
\end{center}
\end{figure}

As shown in Figure~\ref{fig:size-comparison}, the second crawling iteration produced larger web corpora, showing the increase of content on the web. Compared to the size of the CLASSLA-web 1.0 corpora, the web corpora in the 2.0 corpus collection comprise 57\% more words and 46\% more texts, that is, roughly half as many more words and texts than in the first version. Interestingly, web growth is uneven across South Slavic webs -- the Slovenian and Croatian webs exhibit modest increase of size, with a 30\% change in size in terms of number of words, while the size of Bulgarian and Montenegrin web corpora nearly doubled in the second crawling iteration. These differences in sizes reflect the high growth of content on the web rather than any differences in the crawling methodology, as both the CLASSLA-web 1.0 and CLASSLA-web 2.0 corpora have been produced with practically the same pipeline.


\subsection{Small Overlap Between the Two Crawls}
\label{sec:content-overlap}

A comparison of CLASSLA-web 1.0 and 2.0 shows that the new edition of the web corpora contains up to twice as many texts as version 1.0. In this section, we examine the degree of overlap between the 1.0 and 2.0 versions of the CLASSLA-web corpora. This analysis aims to determine whether the growth in corpus size of the 2.0 version can be partially attributed to the inclusion of the majority of the 1.0 corpora within the 2.0 corpora. Additionally, this investigation provides us with insights into the degree of change in the content of South Slavic webs over a short period between the two observed crawls.

We measure content overlap by identifying near-duplicated texts using MinHash over word 4-grams, with a similarity threshold set to 0.7, defined via manual validation. Surprisingly, despite the fact that only two years have passed between the two crawls, there is a small content overlap -- only around 20\% of texts from version 1.0 are present in the CLASSLA-web 2.0 corpora.\footnote{Detailed information indicating which texts in version 2.0 are near duplicates of specific texts in version 1.0 is provided at \url{https://github.com/clarinsi/classla-web/tree/main/more-info/duplicates-with-1.0}.}

In Table~\ref{tab:unique-texts}, we report the share of texts unique to each crawling iteration. On average, 82\% of texts (comprising approximately 81\% of words) in CLASSLA-web 2.0 corpora are new relative to the CLASSLA-web 1.0 corpora. These findings indicate a rapid turnover of web content: most of the content present in version 1.0 has disappeared in two years, and the majority of the content in version 2.0 has appeared recently.

Taking into account also changes in corpora sizes, shown in Figure \ref{fig:size-comparison}, Bulgarian (bg) and Montenegrin (cnr) are the fastest-changing and expanding webs. The CLASSLA-web 2.0 corpora from these two national webs nearly doubled in size compared to the 1.0 versions, while retaining only 11--14\% of content that was already present in the 1.0 corpora. In contrast, Croatian web changed the least, with 24\% of texts in version 2.0 already being present in version 1.0.

\begin{table}[!ht]
\begin{center}
\begin{tabularx}{\columnwidth}{|l|X|X|}
    \hline
        corpus & \% unique texts in CLASSLA-web 2.0 & \% unique texts in CLASSLA-web 1.0 \\ \hline
        bs & 79.41\% & 73.52\% \\ \hline
        bg & 85.79\% & 72.08\% \\ \hline
        hr & 76.08\% & 73.85\% \\ \hline
        mk & 79.94\% & 71.42\% \\ \hline
        cnr & 88.68\% & 77.32\% \\ \hline
        sr & 79.67\% & 70.63\% \\ \hline
        sl & 80.60\% & 77.09\% \\ \hline
        average & 81.77\% & 73.09\% \\ \hline
    \end{tabularx}
    \caption{Percentage of texts in CLASSLA-web 1.0 and CLASSLA-web 2.0 corpora that are unique to that crawling iteration, that is, they do not appear in the other web corpus version.}
    \label{tab:unique-texts}
 \end{center}
\end{table}

\begin{figure}[!ht]
\begin{center}
\includegraphics[width=\columnwidth]{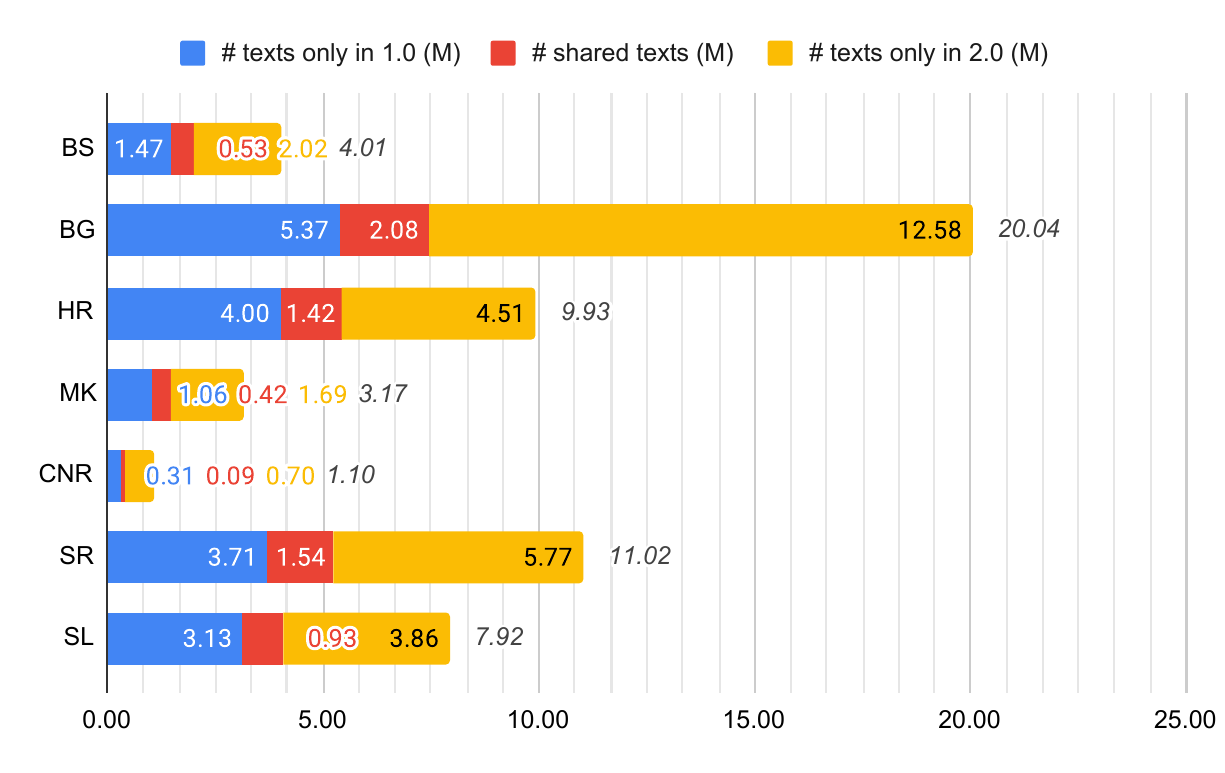}
\caption{Number of texts that are unique in CLASSLA-web 1.0 or 2.0 versions, number of texts that are shared between the two versions, and a total number when the two corpora are merged.}
\label{fig:corpora-overlap}
\end{center}
\end{figure}

The small content overlap also suggests that through iterative crawling over a two-year period, it is possible to collect large quantities of texts. Figure \ref{fig:corpora-overlap} shows the combined size of CLASSLA-web 1.0 and CLASSLA-web 2.0, with texts from version 1.0 excluded from version 2.0. The total size of the merged CLASSLA-web corpora is 57 million texts, with the Bulgarian corpus amounting to 20 million texts. These results demonstrate the effectiveness of iterative crawling for collecting large-scale text resources for less-resourced languages, and provide a strong motivation to continue with biennial crawling efforts.

\subsubsection{URL Overlap as a Quick Insight into Content Overlap}
\label{sec:url-overlap-calculation}

\begin{figure}[!ht]
\begin{center}
\includegraphics[width=\columnwidth]{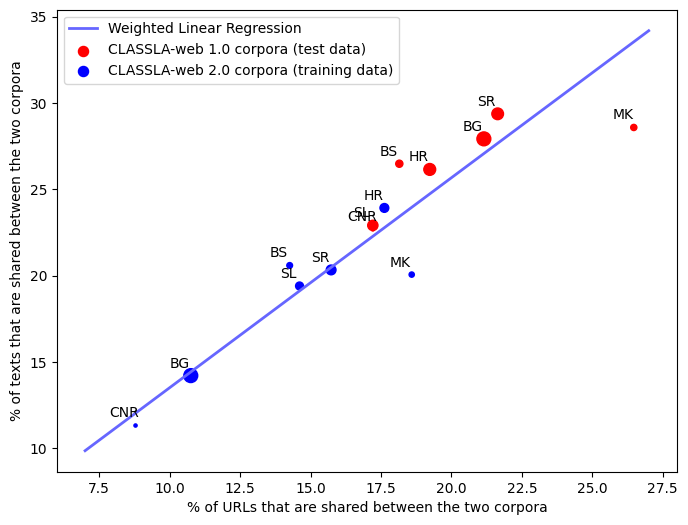}
\caption{Correlation between percentage of shared URLs and shared texts between CLASSLA-web 1.0 and CLASSLA-web 2.0 corpora, and a weighted linear regression predicting percentage of shared texts based on percentage of shared URLs.}
\label{fig:linear-regression-plot}
\end{center}
\end{figure}

In the previous section, we assessed the extent of overlapping content between two corpora using the MinHash algorithm to detect near-duplicate texts. Since this method is computationally intensive, we explore faster alternatives for gaining initial insights into content overlap. One potential indicator of the degree of content overlap is the overlap of URLs of web pages from which the texts were collected, as this information is included in the metadata for each text. Figure \ref{fig:linear-regression-plot} shows a correlation between the percentage of shared URLs and the percentage of shared texts across both corpora. This relationship, based on seven pairs of national CLASSLA-web 1.0 and 2.0 corpora (e.g., CLASSLA-web.hr 1.0 and CLASSLA-web.hr 2.0), is strong, with a Pearson correlation coefficient of 0.908 and a p-value of 0.0047.

As shown in Figure \ref{fig:linear-regression-plot}, this correlation can be modelled using a weighted linear regression, implemented in the scikit-learn Python library\footnote{\url{https://scikit-learn.org/stable/modules/generated/sklearn.linear_model.LinearRegression.html}}. The model was trained on URL and text overlap data from the seven CLASSLA-web 2.0 corpora, specifically their overlap with the corresponding CLASSLA-web 1.0 corpora. The weights were based on the normalised sizes of the CLASSLA-web 2.0 corpora. When tested on CLASSLA-web 1.0 data (i.e., the percentage of overlapping texts and URLs with version 2.0), the model's prediction errors are 0.4--5 percentage points, and below 1.7 percentage points on larger corpora, namely Bulgarian, Croatian, Slovenian and Serbian. On version 2.0, the error ranges from 0.15 to 4 points, with deviations of 1 point or less for the larger corpora.

\begin{equation}
textOverlap = 1.347 + 1.216 \cdot URLOverlap
\label{eq:weighted-regression-function}
\end{equation}

Equation \ref{eq:weighted-regression-function} shows the function derived from the regression model, which can be used to estimate the likely percentage (0-100) of shared texts between two corpora (\textit{textOverlap}) based on the percentage of overlapping URLs (\textit{URLOverlap}). This function offers a rough but useful approximation of content overlap between large web corpora, enabling quick comparisons when detecting near-duplicates directly would be too time-consuming or computationally expensive.

\subsection{Degradation of Web Content}
\label{sec:degradation-of-content}

\begin{figure}[!ht]
\begin{center}
\includegraphics[width=\columnwidth]{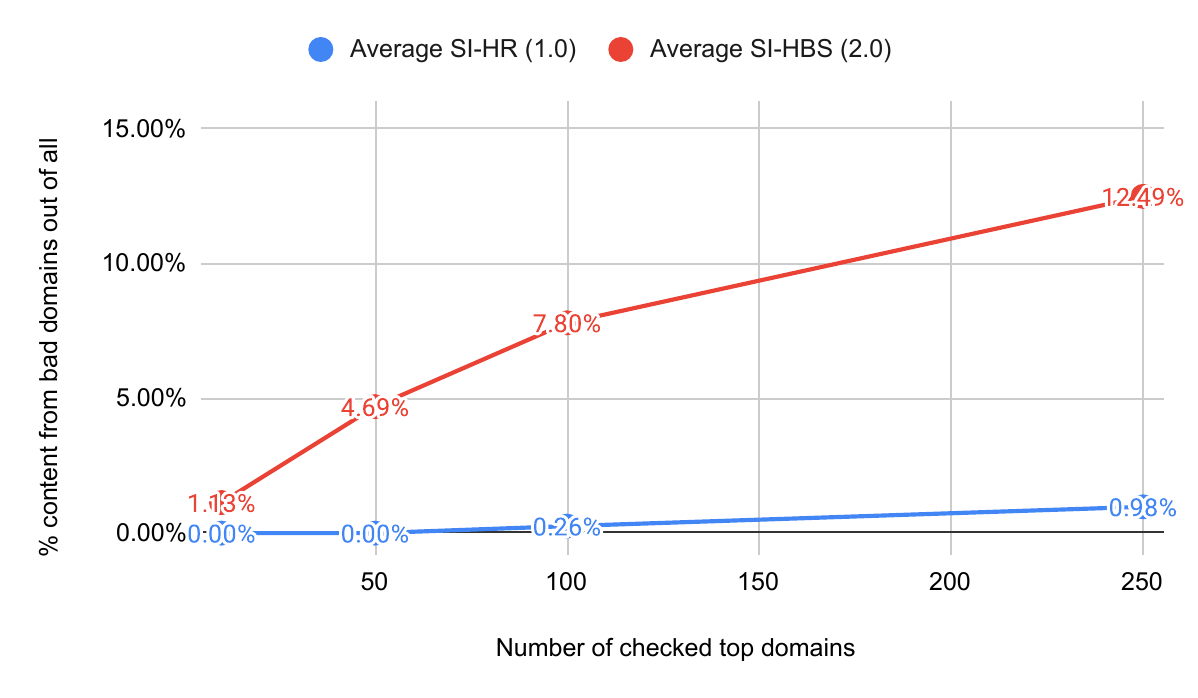}
\caption{Percentage of content (relative to the total corpus content) originating from bad domains among the most frequent web domains in manually inspected Western South Slavic CLASSLA-web 1.0 and CLASSLA-web 2.0 corpora.}
\label{fig:bad-domains}
\end{center}
\end{figure}

While the comparison between versions 1.0 and 2.0 of the CLASSLA-web corpora revealed substantial gains in text content, manual inspection of the most frequent domains also showed a notable increase in low-quality or automatically generated content.\footnote{Lists of domains from the CLASSLA-web corpora that have been manually verified or identified as low quality are provided at \url{https://github.com/clarinsi/classla-web/tree/main/urls}.} As shown in Figure \ref{fig:bad-domains}, a manual review of the Western South Slavic CLASSLA-web 2.0 corpora found that many of the top 250 domains were problematic, accounting for 15\% of the total number of texts before their removal.

In contrast, this issue was much less prominent in the CLASSLA-web 1.0 corpora, crawled in 2021 and 2022. There, low-quality domains rarely appeared among the top 250 domains, and manual inspection led to the removal of only 1\% of texts -- 15 times less than in the 2.0 version, which was crawled in 2024. The likely cause of this surge in low-quality content is the growing availability of automated text generation tools. These findings highlight the importance of manually reviewing the most frequent domains as a necessary step in preparing high-quality web corpora.

\section{Conclusion}
\label{sec:conclusion}

In this paper, we present the CLASSLA-web 2.0 corpus collection for South Slavic languages, a result of an ongoing biennial web crawling infrastructure. Although the corpora were collected from the same national domains just two years after the CLASSLA-web 1.0 collection \citep{ljubevsic2024classlaweb}, the new datasets are significantly larger and contain mostly new content. On average, the 2.0 versions include around 50\% more words and texts than the 1.0 versions. In total, the CLASSLA-web 2.0 corpora consist of 17 billion words and 38 million texts in seven South Slavic languages. The corpora are freely available in JSONL and VERT formats from the CLARIN.SI repository \citeplanguageresource{classlaweb2-repository}\footnote{\url{http://hdl.handle.net/11356/2079}} and can be queried via the CLARIN.SI concordancers\footnote{\url{https://www.clarin.si/ske/}}. More information on corpora development and accessibility is available on a website dedicated to the CLASSLA-web collections (\url{https://clarinsi.github.io/classla-web/}).

In addition to providing metadata on the collected texts, the corpora are automatically annotated with genre, topic, and linguistic information to support various types of linguistic analyses. The genre and topic annotations offer valuable insights into the characteristics of web content and reveal interesting similarities and differences across the South Slavic web corpora. Our analysis shows that all South Slavic web corpora are predominantly composed of news content, which is largely focused on five main topics: ``\textit{Sport}'', ``\textit{Politics}'', ``\textit{Economy, Business and Finance}'', ``\textit{Arts, Culture, Entertainment and Media}'', and ``\textit{Human Interest}''.

Comparison with the previous CLASSLA-web 1.0 versions showed that approximately 80\% of the content in each version is unique, indicating that only about one-fifth of the CLASSLA-web 2.0 content overlaps with the CLASSLA-web 1.0 release. These results demonstrate that iterative crawling of top-level national domains is a highly effective approach for collecting large-scale text data for less-resourced languages. 

Precise measurement of overlapping content requires identifying near-duplicate texts, which is computationally expensive and time-consuming for massive corpora. As an additional contribution, we propose using the percentage of shared URLs as a rough estimate of the amount of overlapping content between web corpora. Based on overlap calculations for seven pairs of 1.0 and 2.0 versions of the CLASSLA-web corpora, we find a strong correlation between URL and text overlap. Building on this, we propose a simple weighted linear model (Equation~\ref{eq:weighted-regression-function}) that provides quick estimates of content overlap when near-duplicate detection is not feasible.

Since the CLASSLA-web 2.0 corpora were developed using the same pipeline as the previous version, CLASSLA-web 1.0, this enables longitudinal analyses of changes in South Slavic web content between the two crawls, that is, between 2021 and 2024. Most notably, there has been a significant increase in low-quality or automatically generated content. Manual inspection of the most frequent domains in CLASSLA-web 2.0 reveals that low-quality domains account for a much larger share of both texts and words compared to version 1.0 and appear far more often among the top-ranked domains. Specifically, an inspection of the top 250 domains identified a large number of bad domains, which contributed 15\% of the total number of texts prior to their removal from the cleaned corpora. These findings underline the growing importance of a manual validation of high-frequency domains to maintain web corpus quality, especially in light of the widespread availability of tools for automatic large-scale text generation.

To conclude, despite the growing pains related to maintaining high-quality web corpora that would be free from automatically-generated texts, web crawling remains a highly effective method for collecting large-scale text resources for less-resourced languages such as South Slavic languages. Due to the rapid turnover of online content, repeating the crawl every two years can more than double the size of the total collected corpora and our results show small content overlap between versions collected two years apart. Thanks to their size and recency, the CLASSLA-web 2.0 corpora have already attracted interest within the South Slavic research and language technology development communities. The corpora have been included in the pretraining data for Slovenian open-source large language models \citep{vrevs2024generative} and introduced to linguists through recent editions of the CLASSLA-Express workshops that focus on the use of web corpora and concordancers for linguistic analyses \citep{ljubevsic2024classlaexpress}. Highly motivated by the community’s uptake of this new web corpus collection, we will continue our biennial web crawling efforts to provide increasingly bigger and up-to-date web corpora for South Slavic languages.


\section{Ethical Considerations and Limitations}

We are aware that using web-collected data can raise questions regarding intellectual property and the privacy rights of the original authors. The web crawling pipeline is designed to avoid collecting sensitive data by only retrieving texts that are freely accessible. Nevertheless, we recognize that some texts in the datasets may still have been included without the authors’ explicit consent. To address this, the CLASSLA-web corpora are published with a notice informing authors that their texts can be removed from the corpora upon request.

\section{Acknowledgements}

The research presented in this paper was conducted within the research project ``Basic Research for the Development of Spoken Language Resources and Speech Technologies for the Slovenian Language'' (J7-4642), the research project ``Large Language Models for Digital Humanities'' (GC-0002), the research project ``Embeddings-based techniques for Media Monitoring Applications'' (L2-50070, co-funded by the Kliping d.o.o. agency), the Research Infrastructure DARIAH-SI (I0-E007), and within the research programme ``Language resources and technologies for Slovene'' (P6-0411), all funded by the Slovenian Research and Innovation Agency (ARIS).

This research was also supported by LLMs4EU, co-funded by the Digital Europe Programme under GA 101198470. This project is co-funded by the European Union. Views and opinions expressed are however those of the author(s) only and do not necessarily reflect those of the European Union or the European Commission. Neither the European Union nor the granting authority can be held responsible for them.

We would like to thank the \href{https://www.clarin.si/info/k-centre/}{CLASSLA knowledge centre for South Slavic languages} and the Slovenian \href{https://www.clarin.si/info/about/}{CLARIN.SI infrastructure} for their valuable support.

\section{Bibliographical References}\label{sec:reference}

\bibliographystyle{lrec2026-natbib}
\bibliography{references}

\section{Language Resource References}
\label{lr:ref}
\bibliographystylelanguageresource{lrec2026-natbib}
\bibliographylanguageresource{languageresource}


\label{sec:appendix}

\end{document}